\documentclass{article}
\usepackage{spconf,amsmath,graphicx}
\usepackage{multirow}
\usepackage{siunitx}
\usepackage{tabularx,booktabs}
\usepackage{xcolor}
\usepackage{comment}
\usepackage{multirow}
\usepackage{booktabs}
\usepackage{tikz}
\usepackage{makecell}

\usepackage{amssymb}
\usepackage{pifont}
\newcommand{\cmark}{\ding{51}}%
\newcommand{\xmark}{\ding{55}}
\usepackage{color}
\usepackage{fancyhdr,graphicx,amsmath,amssymb}
\usepackage{arydshln}
\usepackage{enumitem}
\usepackage{verbatim}
\usepackage{pifont}
\usepackage{footmisc}
\usepackage{graphicx}
\usepackage{booktabs}
\usepackage{float}
\usepackage{amsmath}

\usepackage{hyperref}
\hypersetup{
  colorlinks   = true, 
  urlcolor     = blue, 
  linkcolor    = red, 
  citecolor   = green 
}
\title{Open-Set Automatic Target Recognition}
\name{Bardia Safaei$^{1}$, Vibashan VS$^{1}$,  Celso M. de Melo$^{2}$, Shuowen Hu$^{2}$, and Vishal M. Patel$^{1}$}
\address{$^{1}$ Johns Hopkins University, Baltimore, MD, USA\\
$^{2}$ DEVCOM Army Research Laboratory, Adelphi, MD, USA\\
\texttt{\{bsafaei1, vvishnu2, vpatel36\}@jhu.edu}\\
\texttt{\{celso.m.demelo.civ\}@army.mil}}
\begin{document}
\topmargin=0mm
\ninept
\maketitle
\begin{abstract}
Automatic Target Recognition (ATR) is a category of computer vision algorithms which attempts to recognize targets on data obtained from different sensors. ATR algorithms are extensively used in real-world scenarios such as military and surveillance applications. Existing ATR algorithms are developed for traditional closed-set methods where training and testing have the same class distribution. Thus, these algorithms have not been robust to unknown classes not seen during the training phase, limiting their utility in real-world applications. To this end, we propose an Open-set Automatic Target Recognition framework where we enable open-set recognition capability for ATR algorithms. In addition, we introduce a plugin Category-aware Binary Classifier (CBC) module to effectively tackle unknown classes seen during inference. The proposed CBC module can be easily integrated with any existing ATR algorithms and can be trained in an end-to-end manner. Experimental results show that the proposed approach outperforms many open-set methods on the DSIAC and CIFAR-10 datasets. To the best of our knowledge, this is the first work to address the open-set classification problem for ATR algorithms. Source code is available at:
\url{https://github.com/bardisafa/Open-set-ATR}.
\end{abstract}
\vspace{- 1.0 em}
\section{Introduction}
\label{sec:intro}

Automatic Target Recognition (ATR) algorithms process information acquired from multiple  sensors (i.e. visible and infrared) to recognize the targets appearing in the scene \cite{dudgeon1993overview, rogers1995neural, Patel_ATRICIP2010, Patel_ATR_SRC}. These ATR algorithms are capable of detecting targets at different scales and ranges (Range:1000m - 5000m) \cite{vs2022meta}, which are often not recognized by the naked eye. Further, these algorithms can effectively remove the human intervention from the process of target acquisition and recognition  \cite{bhanu1986automatic}, making them a successful automatic recognition system. Thus, ATR systems are extensively used in different commercial and military applications \cite{bhanu1986automatic}. An ATR algorithm consists of two major components; detection and classification. The detection component generally involves a computationally simple region proposal pipeline where target proposals are generated from motion or by eliminating surrounding clutter \cite{kazemi2019deep}. The classification component involves feature extraction and pattern recognition for classifying target categories.

The rise in Deep Neural Networks (DNNs) \cite{krizhevsky2017imagenet} has drastically improved the performance of computer vision tasks such as image classification, image segmentation and object detection \cite{krizhevsky2017imagenet, ren2015faster, felzenszwalb2004efficient}. Further, the increase in computational resources and datasets have enabled these deep networks to be deployable in real-world applications. Recently, there has been significant interest in improving ATR algorithms using DNNs \cite{kazemi2019deep, vs2022meta} where the DNN frameworks are employed to improve detection and classification components of the ATR algorithms \cite{ren2015faster, he2016deep}. Unlike the traditional ATR methods, these deep learning frameworks eliminate the need for problem-specific discriminative features designed by humans. One of the major challenges in existing DNN frameworks is that they are based on the closed-world assumption; the model assumes only the classes seen during training will appear in the real-world. This assumption limits the utility of existing DNN-based ATR systems in actual applications. Therefore, it is important to develop a DNN-based ATR algorithm that detects unknown classes not seen during training.

\begin{figure}[t!]
    \begin{center}
        \includegraphics[width=\linewidth]{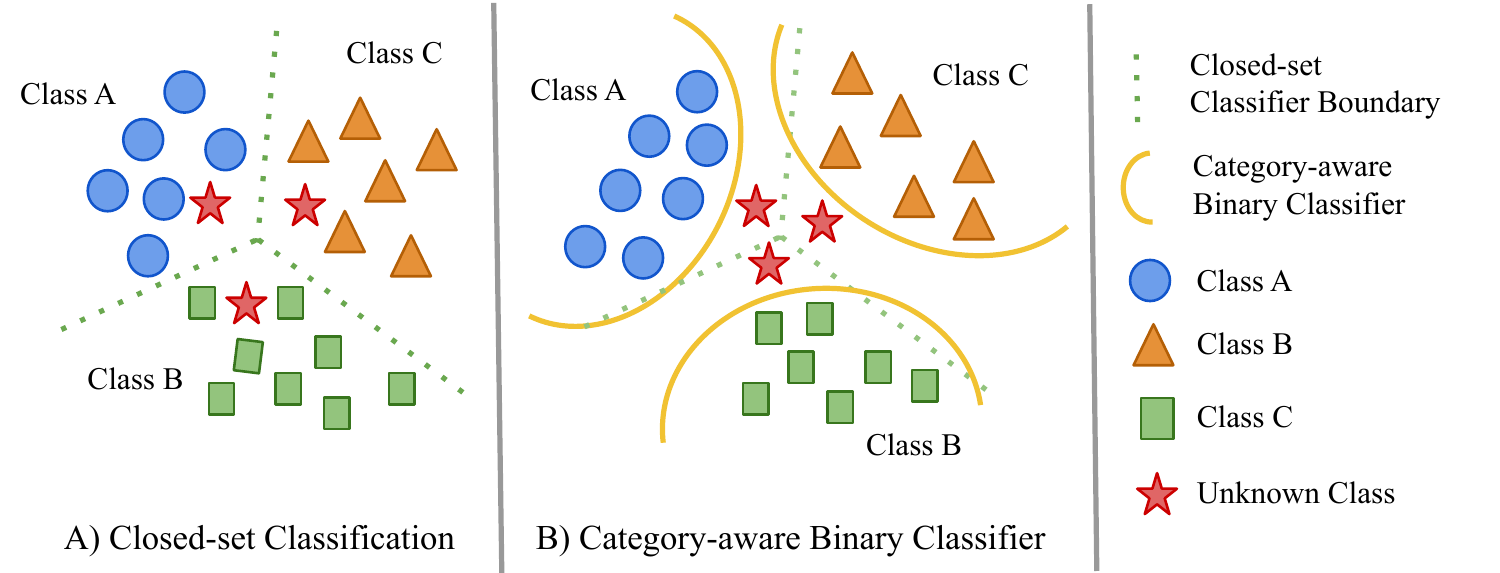}
    \end{center}
    \vskip -15.0pt
    \caption{ A) A closed-set classifier generates a multi-class decision boundary to distinguish between known samples. Given unknown samples, the closed-set classifier classifies them as known samples with high confidence because the closed-set classifier is never trained for unknown samples. Therefore, relying on a multi-class decision boundary to detect unknown samples is not an optimal solution. B) The proposed Category-aware Binary Classifier learns a category-aware representation resulting in a compact category-aware decision boundary. Hence, combining multi-class and category-ware decision boundaries effectively classifies known and unknown samples. }
    \label{fig:intro} 
    \vskip -15.0pt
\end{figure}

\begin{figure*}[t!]
    \begin{center}
        \includegraphics[width=.9\linewidth]{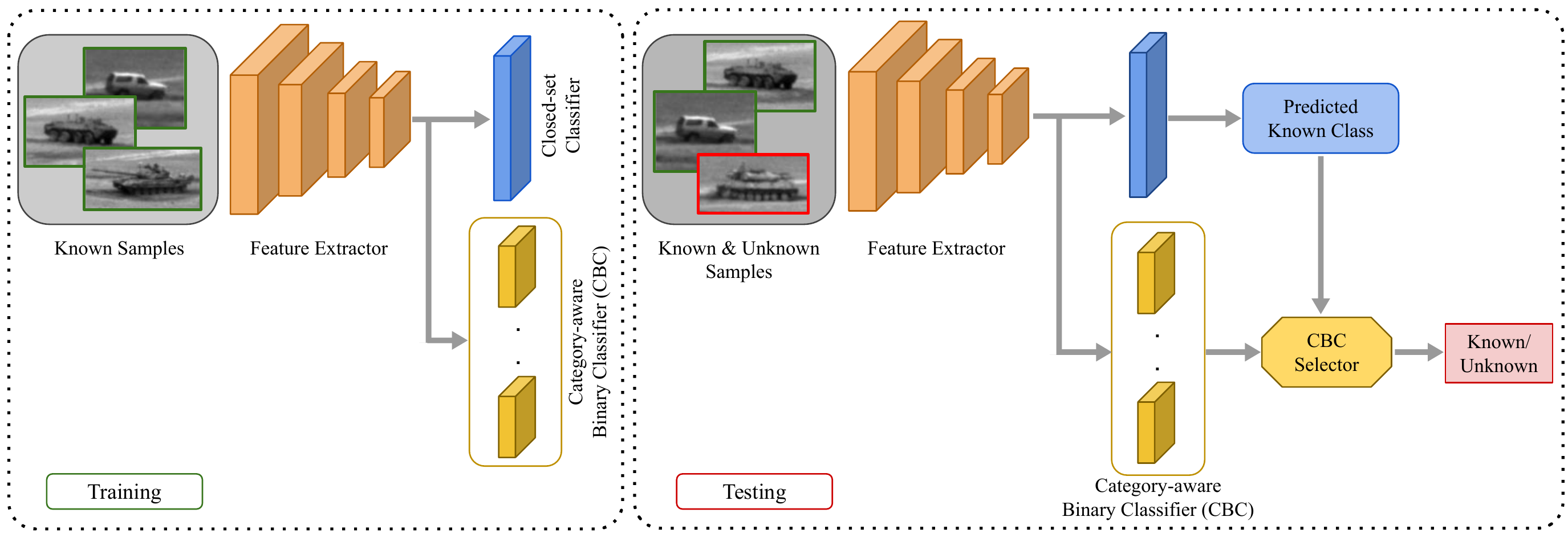}
    \end{center}
    \vskip -17.0pt 
    \caption{Overview of our proposed approach for open-set automatic target recognition. \textbf{Training:} For a given batch of images, the encoder network extracts features and the closed-set classifier classifies the features into targets. In addition, the extracted features are fed into a category-aware binary classifier where each binary classifier learns the decision boundary to classify whether a given sample belongs to that category or not. \textbf{Testing:} For a given image, the closed classifier predicts the nearest known class and its corresponding binary classifier score is used to decide whether the given sample belongs to that category or not.} 
    \label{fig:archt} 
    \vskip -15.0pt
\end{figure*}

Open-set recognition is a problem of handling `unknown' classes that are not seen during training, whereas traditional closed-set classifiers assume that only known classes appear during testing \cite{geng2020recent, vaze2021open}. Existing open-set recognition methods are generally classified into generative and discriminative methods. G-OpenMax \cite{ge2017generative} is a generative method which trains a model with synthesized unknown data. However,  these methods cannot be applied to natural images other than hand-written characters due to the difficulty of generative modelling. OpenMax \cite{bendale2016towards} is a discriminative method which trains a model with softmax cross-entropy classification loss for closed-set samples and unknown classes are detected by applying a threshold on predicted probabilities or logit scores \cite{bendale2016towards}. Outlier (also called anomaly or novelty) detection can be incorporated into the concept of open-set classification as an unknown detector. 
Some of the generic methods for outlier detection are one-class extensions of discriminative models such as SVM \cite{manevitz2001one} or forests \cite{liu2008isolation} and subspace methods \cite{ringberg2007sensitivity}. However, outlier detectors are not open-set classifiers by themselves because they have no discriminative power within known classes and it cannot be used to build a general purpose open-set classifier. Therefore, extending discriminative open-set recognition methods is a promising approach towards solving the open-set ATR problem. 

In this work, we employ ResNet 18 \cite{he2016deep} as the base network for ATR closed-set classification. Our experiments have shown that for open-set classification, employing OpenMax \cite{bendale2016towards} or SoftMax \cite{Neal_2018_ECCV} for classifying targets in short ranges (0m-1000m) works reasonably well. However, the open-set classification performance drops as the input data range increases (see in Table \ref{open-acc}). This is due to the fact that as the range increases the target size decreases and the network finds it difficult even to learn discriminative known class features resulting in poor performance. To overcome this issue, we propose a novel plugin Category-aware Binary Classifier (CBC) module which effectively tackles the unknown categories seen during inference at various ranges. In CBC, for each category, a binary classifier is trained using the samples of the corresponding category as known samples. The remaining samples from other categories are treated as unknown samples. In this way, each binary classifier learns category-aware decision boundaries, which decide whether a given sample belongs to that category or not. In the testing phase, the closed-set classifier identifies the nearest known class and the corresponding binary classifier's score is leveraged to decide whether it is a known or unknown sample. Hence, combining multi-class and category-ware decision boundaries effectively classifies known and unknown samples at various ranges making it more suitable for ATR (see Fig. 1).  This paper makes the following contributions:
\begin{itemize}[topsep=0pt,noitemsep,leftmargin=*]  
    \item To the best of our knowledge, this is the first work to consider an open-set recognition problem for automatic target recognition. 
    \item We propose a plugin Category-aware Binary Classifier (CBC) module to effectively tackle unknown classes seen during inference at various ranges.
    \item We consider the DSIAC and CIFAR-10 benchmark datasets for experimental analysis and show that the proposed method outperforms many open-set methods.
\end{itemize}

\begin{table*}[!t]
\renewcommand{\arraystretch}{1.4}

\centering
\begin{tabular}{c|cc|cc|cc|cc}
\hline \hline
\multirow{2}{*}{Method} & \multicolumn{2}{c|}{Range 1000}        & \multicolumn{2}{c|}{Range 2000}        & \multicolumn{2}{c|}{Range 3000}        & \multicolumn{2}{c}{Combined}          \\ \cline{2-9} 
                        & \multicolumn{1}{c|}{Visible} & Thermal & \multicolumn{1}{c|}{Visible} & Thermal & \multicolumn{1}{c|}{Visible} & Thermal & \multicolumn{1}{c|}{Visible} & Thermal \\ \hline
ConfLabel \cite{lee2021open}                     & \multicolumn{1}{c|}{72.05$\pm$6.9}        &  74.29$\pm$3.9       & \multicolumn{1}{c|}{70.69$\pm$6.3}        &72.88$\pm$10.0         & \multicolumn{1}{c|}{69.27$\pm$9.5}        &  71.39$\pm$ 5.9        & \multicolumn{1}{c|}{68.78$\pm$5.8}        & 63.60$\pm$7.2        \\ 
MLS \cite{vaze2021open}                 & \multicolumn{1}{c|}{95.71$\pm$4.3}        &94.59$\pm$1.0         & \multicolumn{1}{c|}{90.83$\pm$2.2}        &93.50$\pm$5.4         & \multicolumn{1}{c|}{87.68$\pm$5.1}        &83.73$\pm$10.2         & \multicolumn{1}{c|}{91.74$\pm$2.4}        &88.81$\pm$1.7          \\ 
SoftMax \cite{Neal_2018_ECCV}                 & \multicolumn{1}{c|}{96.10$\pm$3.9}        &94.33$\pm$0.9         & \multicolumn{1}{c|}{89.41$\pm$4.8}        &94.94$\pm$3.7         & \multicolumn{1}{c|}{89.71$\pm$5.8}        &86.07$\pm$7.1         & \multicolumn{1}{c|}{94.66$\pm$0.9}        &92.04$\pm$1.2         \\ 
OpenMax \cite{bendale2016towards}                   & \multicolumn{1}{c|}{94.74$\pm$3.9}        &93.85$\pm$0.6         & \multicolumn{1}{c|}{90.04$\pm$4.1}        &93.41$\pm$4.1         & \multicolumn{1}{c|}{91.57$\pm$7.7}        &87.00$\pm$6.2         & \multicolumn{1}{c|}{93.92$\pm$1.8}        &90.80$\pm$1.2         \\ \hdashline
Ours                    & \multicolumn{1}{c|}{\textbf{98.61}$\pm$1.3}        &\textbf{96.27}$\pm$1.1         & \multicolumn{1}{c|}{\textbf{93.28}$\pm$1.7}        &\textbf{96.17}$\pm$1.7         & \multicolumn{1}{c|}{\textbf{92.03}$\pm$7.3}        &\textbf{87.52}$\pm$8.6         & \multicolumn{1}{c|}{\textbf{94.89}$\pm$1.4}        &\textbf{92.93}$\pm$2.3         \\ \hline \hline
\end{tabular}
\caption{Open-set identification results corresponding to different approaches on the DSIAC dataset. The averaged AUROC(\%) results are reported after repeating each experiment for three times with randomly selected
known/unknown class splits.}
\vskip -10.0pt
\label{vis-ther}
\end{table*}

\vspace{- 1.0 em}
\section{Method}
Let us define the dataset as $\mathcal{D}= \mathcal{D}_{train}\cup\mathcal{D}_{test}$. We split $\mathcal{D}_{train}$ into \textit{knowns}, $\mathcal{K}_{train}$, and \textit{unknowns}, $\mathcal{U}_{train}$. We only use $\mathcal{K}_{train}=\left\{ \left( X^{i},y^{i} \right) \right\}_{i=1}^{N_{train}}$ to develop our open-set recognition model where $y^i\in \left\{ j \right\}_{j=1}^{N_k}$, $N_{train}$ is the size of training data, and $N_k$ is the number of known classes. We then test the model on $\mathcal{D}_{test}= \mathcal{K}_{test}\cup\mathcal{U}_{test}$. Two distinct branches make up our training phase. A closed-set classifier is trained in one branch, and category-aware binary classifiers (CBC) for open-set identification are trained in the other. We train the model in an end-to-end manner. 
\label{sec:format}
\subsection{Closed-set training}
For open-set ATR, we utilize a closed-set classifier with a feature extractor ($F$) and a fully connected layer ($C$) to classify known samples. We employ the Cross Entropy (CE) loss to train a closed-set classifier on $\mathcal{K}_{train}$ and update the parameters of $F$ and $C$. Given an input image $ \left( X,y \right)\in\mathcal{K}_{train}$, the CE loss is defined as:  
\setlength{\belowdisplayskip}{0pt} \setlength{\belowdisplayshortskip}{0pt}
\setlength{\abovedisplayskip}{0pt} \setlength{\abovedisplayshortskip}{0pt}
\begin{equation}
    \mathcal{L}_{ce} (X, y)=-\sum_{i=1}^{N_k}\mathbb{I}_{y}(i)\log(p_{C}^{i}),
\end{equation}
where $\mathbb{I}_{y}(i)$ is the indicator function of label $y$ and $p_{C}^{i}=\sigma\left( C\left( Z \right) \right)$ is the predicted probability of $X$ belonging to the $i$-th class. Here $Z=F(X)$ is the feature vector and $\sigma$ is the softmax activation function. Further, we perform entropy minimization on known samples to regularize the closed-set classifier for various ranges:
\begin{equation}
    \mathcal{L}_{ent} (X, y)= -\sum_{i=1}^{N_{k}} p_{C}^{i}\log\left( p_{C}^{i} \right).
\end{equation}
\subsection{Open-set training}
To enable open-set recognition, we propose a category-aware binary classifier where each binary classifier learns a decision boundary to classify whether a given sample belongs to that category or not. To train a binary classifier for a category, we typically consider all the samples from that category to be positive and the rest as negative. However, in practice, simply training binary classifiers with lots of negative samples will skew the decision boundaries towards negative samples. To address this problem, we leverage the idea of \textit{hard sample selection} \cite{xie2022learning, saito2021ovanet} where for each sample, the positive and the nearest negative-class boundaries are updated. 
We can decrease the negative-class bias in binary classifiers by training them with the following loss, which works better than BCE (see Section \ref{sec:ab_study}):
\begin{equation}
\mathcal{L}_{cbc} (X, y)= -\log(p_{O}^y) - \min_{i\neq y}\log(1-p_{O}^i),
\end{equation}
where $p_{O}^i=\sigma\left( f_i\left( Z \right) \right)$ is the probability of $X$ being classified as positive (known) by $f_i$, and $\left\{ f_i \right\}_{i=1}^{N_{k}}$ are category-aware binary classifiers.
The total loss function for training the model is given as follows:
\begin{equation}
    \mathcal{L}_{total} =  \mathcal{L}_{ce}(X,y)+\mathcal{L}_{cbc}(X,y)+\lambda\mathcal{L}_{ent}(X,y).
\end{equation}

\vspace{- 1.0 em}
\subsection{Testing}

 For a given image $X_{test}$, we first compute the feature map $Z_{test}=F(X_{test})$ and then classify it into one of the known classes, i.e., $\hat{y}_{test}=C(Z_{test}). $ Now we select the corresponding binary classifier in CBC as the open-set detector to see whether the category-aware decision boundary detects $X_{test}$ as unknown or not. Thus, we compare $p_{O}^{\hat{y}_{test}}=\sigma\left( f_{\hat{y}_{test}}\left( Z_{test} \right) \right)$ with a predefined threshold $\gamma$. If $p_{O}^{\hat{y}_{test}}<\gamma$ we categorize the input as an unknown sample and if $p_{O}^{\hat{y}_{test}}>\gamma$ we categorize the input as an known sample. Overview of our proposed training and testing stages is shown in Fig. \ref{fig:archt}. 

\section{Experimental Setup}
\label{sec:pagestyle}
\subsection{Datasets}
\begin{figure*}[t!]
    \begin{center}
        \includegraphics[width=.9\linewidth]{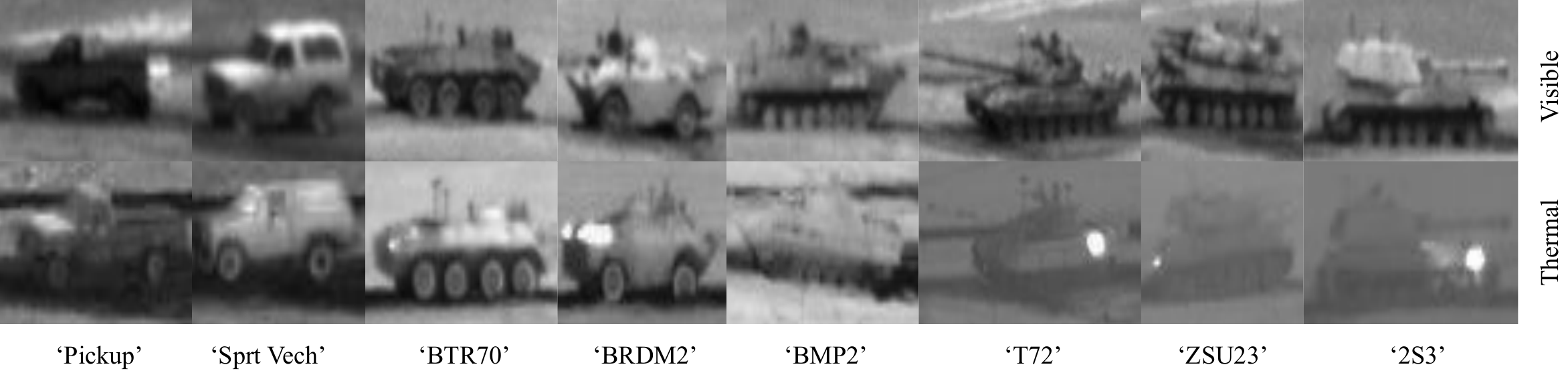}
    \end{center}
    \vskip -15.0pt
    \caption{Sample images for each class of the DSIAC dataset in thermal and visible domains.}
    \label{fig:intro_data} 
    \vskip -10.0pt
\end{figure*}

We conduct our experiments using two publicly available datasets: DSIAC \cite{WinNT} and CIFAR-10 \cite{krizhevsky2009learning}. The DSIAC dataset provides images in two domains, visible and thermal, which include eight classes of civilian and military vehicles, namely 'Pickup', 'Sport vehicle',
‘BTR70’, ‘BRDM2’, ‘BMP2’, ‘T72’, ‘ZSU23’, ‘2S3’. These images were taken at five different ranges, from 1000 to 5000 meters and at intervals of 1000 meters (see Fig.~\ref{fig:intro_data}). To use the dataset, we first crop the target in each image using the bounding-box information provided in the dataset and then resize the target to the size of $224\times224$ while retaining the target ratio. 
Fig. \ref{fig:intro_data} shows sample images from this dataset. We use CIFAR-10, which is a commonly used dataset for image classification tasks, to show that our method can be generalized well to other benchmark datasets. 

\subsection{Implementation Details} 
In all experiments, we train the models for 20 epochs using the cross entropy loss. The \texttt{SGD} optimizer \cite{ruder2016overview} is used with the learning rate of 0.001, the momentum of 0.9, and the weight decay of 0.0005. The weight of entropy loss ($\lambda$) is set to 0.1. The threshold value $\gamma$ is set to 0.9 for the deployment of our method. We use a batch size of 64 for CIFAR-10 and 32 for DSIAC. We use ResNet 18 architecture for the feature extractor and a fully connected layer for the binary classifiers. The implementation is in PyTorch \cite{paszke2019pytorch}, and we utilize an NVIDIA TitanX GPU.

\section{Experiments and Results}
\label{sec:results}

\begin{table}[!t]
\renewcommand{\arraystretch}{1.4}

\resizebox{\columnwidth}{!}{%
\centering
\begin{tabular}{c|cc|cc|cc}
\hline \hline
\multirow{2}{*}{Method} &  \multicolumn{2}{c|}{Range 1000} & \multicolumn{2}{c|}{Range 3000}        & \multicolumn{2}{c}{Combined}          \\ \cline{2-7} 
                        & \multicolumn{1}{c|}{Visible} & Thermal 
                        & \multicolumn{1}{c|}{Visible} & Thermal & \multicolumn{1}{c|}{Visible} & Thermal \\ \hline
ConfLabel \cite{lee2021open}    & \multicolumn{1}{c|}{61.2}        &61.7                    & \multicolumn{1}{c|}{57.3}        &57.5         & \multicolumn{1}{c|}{59.4}        &55.2         \\ 
MLS \cite{vaze2021open}     &  \multicolumn{1}{c|}{82.3}        &73.8            &  \multicolumn{1}{c|}{74.6}        &75.8         & \multicolumn{1}{c|}{74.8}        &72.1         \\ 
SoftMax \cite{Neal_2018_ECCV}   &  \multicolumn{1}{c|}{85.3}        &82.8               &  \multicolumn{1}{c|}{79.7}        &77.7         & \multicolumn{1}{c|}{81.1}        &78.1         \\ 
OpenMax \cite{bendale2016towards}     & \multicolumn{1}{c|}{83.2}        &83.2              & \multicolumn{1}{c|}{\textbf{81.8}}        &81.6         & \multicolumn{1}{c|}{81.6}        &81.4         \\ \hdashline
Ours           & \multicolumn{1}{c|}{\textbf{91.7}}        &\textbf{88.6}                 & \multicolumn{1}{c|}{81.2}        &\textbf{83.3}         & \multicolumn{1}{c|}{\textbf{86.3}}        &\textbf{86.4}         \\ \hline \hline
\end{tabular}}
\caption{Open-set classification results on the DSIAC dataset for ranges 1000, 3000 and combined. We report accuracy (\%) for a 6-class classification setting where we classify 5 ATR targets as 5 knowns and 3 ATR targets as 1 unknown.}
\label{open-acc}
\vskip -15.0pt
\end{table}


We conduct various experiments to show the effectiveness of our method for open-set identification and classification. Open-set identification measures the error of an open-set algorithm in identifying and rejecting unknown samples during testing.
Following the established protocol for evaluating the open-set recognition performance \cite{Neal_2018_ECCV}, we report area under ROC (AUROC) in the open-set identification experiments. AUROC is a calibration-free metric that determines how good the open-set score is without being affected by the chosen threshold. In open-set classification, the model classifies a given sample into one of the known classes or the unknown class which is an $N_k+1$ - class classification problem and we report accuracy in the open-set classification experiments. Open-set classification shows the ability of an open-set classifier to classify known classes in addition to rejecting unknown samples.
For fair comparison, in all our experiments we randomly select some of the classes to be \textit{knowns} and the remaining classes to be \textit{unknowns}. We use the known classes of the training set for training the models, and both known and unknown classes of the testing set for inference. For the DSIAC dataset, 5 classes are randomly chosen as known classes and 3 as unknown classes. On CIFAR-10, the random class splits are 6 known classes and 4 unknown classes. We conduct the experiments on a set of 3 different randomized class splits, and for each split, we run experiments for three times. 

We compare our approach with several existing open-set algorithms, namely SoftMax \cite{Neal_2018_ECCV}, MLS \cite{vaze2021open}, OpenMax \cite{bendale2016towards}, G-OpenMax \cite{ge2017generative}, and ConfLabel \cite{lee2021open}. In SoftMax, the maximum value of the softmax layer is used for open-set recognition while MLS computes the open-set score from the penultimate layer's logits. OpenMax trains a model with an additional unknown class and uses meta-recognition and Extreme Value Theory (EVT) to calibrate the output probabilities and estimate the probability of an input belonging to the unknown class. G-OpenMax generates synthetic unknown samples by Generative Adversarial Networks (GAN) \cite{ge2017generative} to be used during the training. ConfLabel is a method based on the gradients of samples where it leverages gradient-based features to train an unknown detector. 
  

The open-set identification results of different methods on the DSIAC dataset are shown in Table \ref{vis-ther}. We perform the experiments for images in both visible and thermal domains and for various ranges. In the last two columns of the table, we report the performance on the \textit{Combined} dataset, which includes the collection of images in all three ranges. Table \ref{vis-ther} shows that our proposed approach achieves  higher AUROC scores and outperforms other methods across various ranges and for both visible and thermal images. Particularly, in the visible/thermal domains of the \textit{Combined} dataset, we perform better than SoftMax and OpenMax methods by a considerable margin. Moreover, our method obtains the lowest standard deviation overall compared to all other methods making it more robust and reliable. In Table \ref{open-acc}, we report the open-set classification accuracy of our model on the DSIAC images in ranges 1000m, 3000m and the  \textit{Combined} dataset. As shown in Table \ref{open-acc}, our approach outperforms all other methods by a large margin in terms of classification accuracy. Particularly, in visible/thermal domains of the \textit{Combined} dataset, we improve SoftMax and OpenMax methods by 5.2/8.3\% and 4.7/5.0\%, respectively.
To verify the generalization ability of our method, we also conduct an experiment on the CIFAR-10 benchmark dataset. From Table \ref{cifar} we can infer that our proposed method outperforms MLS and G-OpenMax methods by 3.9\% and 6.1\%, respectively showing the generalization capability of our method.


\begin{table}[!t]
\renewcommand{\arraystretch}{1.4}
\centering
\begin{tabular}{cc}
\hline \hline
        & CIFAR-10 \\ \hline
MLS \cite{vaze2021open}    &69.7          \\ 
SoftMax \cite{Neal_2018_ECCV} &67.7          \\ 
G-OpenMax \cite{ge2017generative} &67.5 \\ 
OpenMax \cite{bendale2016towards}&69.5          \\ \hdashline
Ours    &\textbf{73.6}          \\ \hline \hline 
\end{tabular}
\caption{Quantitative results of different approaches on the CIFAR-10 dataset. The averaged AUROC(\%) results are reported.}
\label{cifar}
\vskip -15.0pt
\end{table}

\vspace{- 0.7 em}

\section{Ablation Study}
\label{sec:ab_study}
Table \ref{ablation} shows ablation study for our method on the \textit{Combined} dataset for both thermal and visible images. In row 1, we conduct an experiment without the CBC loss and with the EM loss and in row 2, we conduct an experiment with the CBC loss and without the EM loss. From row 1 and 2, we can observe that the CBC loss helps in learning a more compact representation for the known category better than the EM loss. This shows the effectiveness of the CBC loss over the EM loss. Finally, when we combine both CBC and the EM loss, we get the improved performance of 94.89 and 92.93  for visible and thermal domain, respectively. The overall performance improvement indicates that our CBC module benefits the open-set identification ability of an ATR algorithm by learning better category-aware representations.

\begin{table}[!t]
\renewcommand{\arraystretch}{1.4}
\centering
\begin{tabular}{c|c||cc}
\hline \hline
\multicolumn{1}{c|}{\multirow{2}{*}{CBC}} & \multirow{2}{*}{EM} & \multicolumn{2}{c}{AUROC}             \\ \cline{3-4} 
\multicolumn{1}{c|}{}                                       &                                       & \multicolumn{1}{c|}{Visible} & Thermal \\ \hline                                                   \xmark     &   \cmark                                   &  \multicolumn{1}{c|}{90.71}        &91.60         \\ 
                                                            \cmark & \xmark                                       &  \multicolumn{1}{c|}{94.21}        &  91.88       \\ 
                                                    \cmark         &    \cmark                                   &  \multicolumn{1}{c|}{\textbf{94.89}}        &  \textbf{92.93}       \\ \hline \hline
\end{tabular}
\caption{Ablation study for our open-set recognition algorithm. The impact of the CBC loss and `Entropy minimization' (EM) on performance has been studied. BCE is used when CBC loss is not.}
\label{ablation}
\vskip -10.0pt
\end{table}

\vspace{- 0.5 em}

\section{Conclusion}
\label{sec:conclusion}
In this work, we proposed an effective open-set recognition algorithm for Automatic Target Recognition (ATR). Specifically, we introduced a plugin Category-aware Binary Classifier (CBC)
module that is able to better identify unknown samples by learning compact category-aware decision boundaries. Furthermore, the integration of the proposed method with existing DNN-based ATR systems is straightforward. Our approach outperforms various open-set recognition techniques on different ATR settings, including visible and thermal domains and at different ranges. We also demonstrated that our method's superiority is not restricted to the ATR situation and can perform just as well with other benchmark datasets. In the future, we will expand on this research work to include not only identifying unknown samples but also being able to classify them as novel classes and use them continuously to enhance the model's performance in closed-set and open-set scenarios.

\noindent{\bf{Acknowledgment:}}
Research was sponsored by the Army Research Laboratory and was accomplished under
Cooperative Agreement Number W911NF-23-2-0008. The views and conclusions contained in this document are those of the authors and should not be interpreted as representing the official policies, either expressed or implied, of the Army Research Laboratory or the U.S. Government. The U.S. Government is authorized to reproduce and distribute reprints for Government purposes notwithstanding any copyright notation herein.


\vfill\pagebreak

\bibliographystyle{IEEEbib}
\bibliography{refs}

\end{document}